\def\year{2019}\relax
\documentclass[letterpaper]{article} %DO NOT CHANGE THIS
\usepackage{aaai19}  %Required
\usepackage{times}  %Required
\usepackage{helvet}  %Required
\usepackage{courier}  %Required
\usepackage{url}  %Required
\usepackage{graphicx}  %Required
\usepackage{array}
\usepackage{subfig}
\usepackage{amsmath}

\newcommand{\citet}[1]
{\citeauthor{#1}~\shortcite{#1}}
\newcommand{\citep}{\cite}

\begin{document}
\title{Network Recasting: A Universal Method for Network Architecture Transformation}
\author{Joonsang Yu, Sungbum Kang and Kiyoung Choi\\
Dept. of Electrical and Computer Engineering\\
Neural Processing Research Center (NPRC)\\
Seoul National University, Seoul, Korea\\
\{joonsang.yu, sb05kang\}@dal.snu.ac.kr, kchoi@snu.ac.kr\\
}
\maketitle
\begin{abstract}
This paper proposes network recasting as a general method for network architecture transformation. 
The primary goal of this method is to accelerate the inference process through the transformation, but there can be many other practical applications. 
The method is based on block-wise recasting; it recasts each source block in a pre-trained teacher network to a target block in a student network. 
For the recasting, a target block is trained such that its output activation approximates that of the source block. 
Such a block-by-block recasting in a sequential manner transforms the network architecture while preserving the accuracy. 
This method can be used to transform an arbitrary teacher network type to an arbitrary student network type. 
It can even generate a mixed-architecture network that consists of two or more types of block. 
The network recasting can generate a network with fewer parameters and/or activations, which reduce the inference time significantly. 
Naturally, it can be used for network compression by recasting a trained network into a smaller network of the same type. 
Our experiments show that it outperforms previous compression approaches in terms of actual speedup on a GPU.
\end{abstract}

\section{Introduction} \label{sec:intro}
Deep Neural Networks (DNNs) are widely used for many kinds of recognition and classification tasks because it has outperformed previous machine learning methods in terms of inference accuracy. 
New kinds of DNN architecture have been introduced to achieve even higher accuracy \cite{lin2013network,szegedy2015going,larsson2016fractalnet,he2016deep,zagoruyko2016wide,huang2017densely}, and the networks become deeper and deeper to take the exponential advantage of depth \cite{goodfellow2016deep}. 
To train a deep network, \citet{he2016deep} proposed the residual network (ResNet), which consists of the summation of identity mapping and output of convolutional layers. 
It helps to propagate gradients from top layer to bottom layer, so it can alleviate the vanishing-gradient problem. 
The densely connected network (DenseNet) is also proposed to solve that problem and improve information flow \cite{huang2017densely}. 
DenseNet uses the feature concatenation method instead of summation, so bottom layers can access gradients directly through the concatenation path. 

Deeper network architectures help to achieve higher accuracy, but those have a huge amount of parameters and computation redundancies. 
To design a compact network architecture, the $1\times1$ convolution is added \cite{szegedy2015going,he2016deep,huang2017densely}. 
The additional $1\times1$ convolution reduces the number of channels of output activation. 
The number of parameters and multiplications in the $3\times3$ convolution are also reduced thanks to the $1\times1$ convolution. 
In this reason, bottleneck block in ResNet and dense block in DenseNet use the $1\times1$ convolution for the parameter and multiplication reduction.

\begin{figure}[t]
\centering
	\subfloat{\includegraphics[width=0.5\columnwidth]{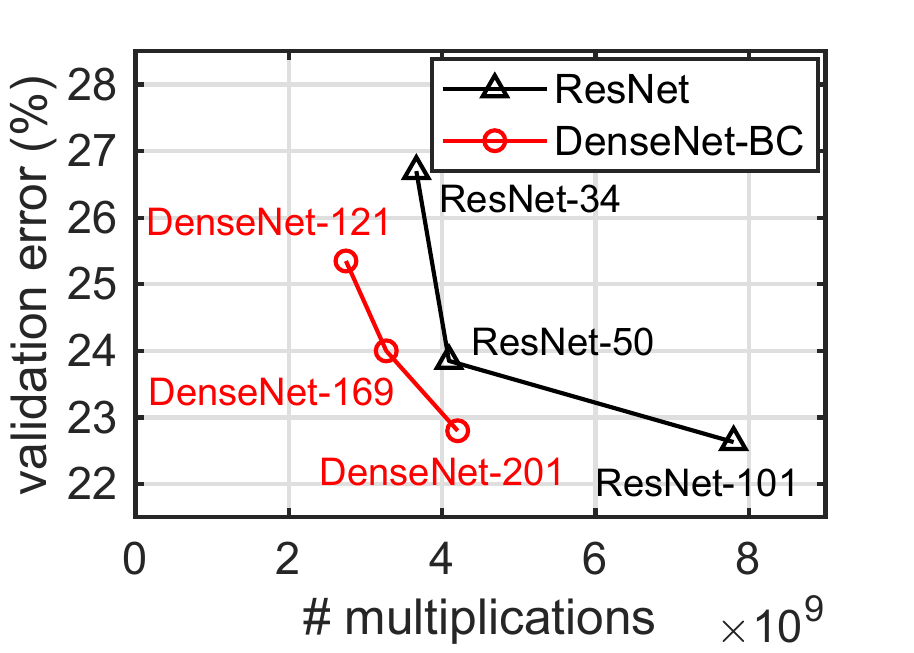}}
	\subfloat{\includegraphics[width=0.5\columnwidth]{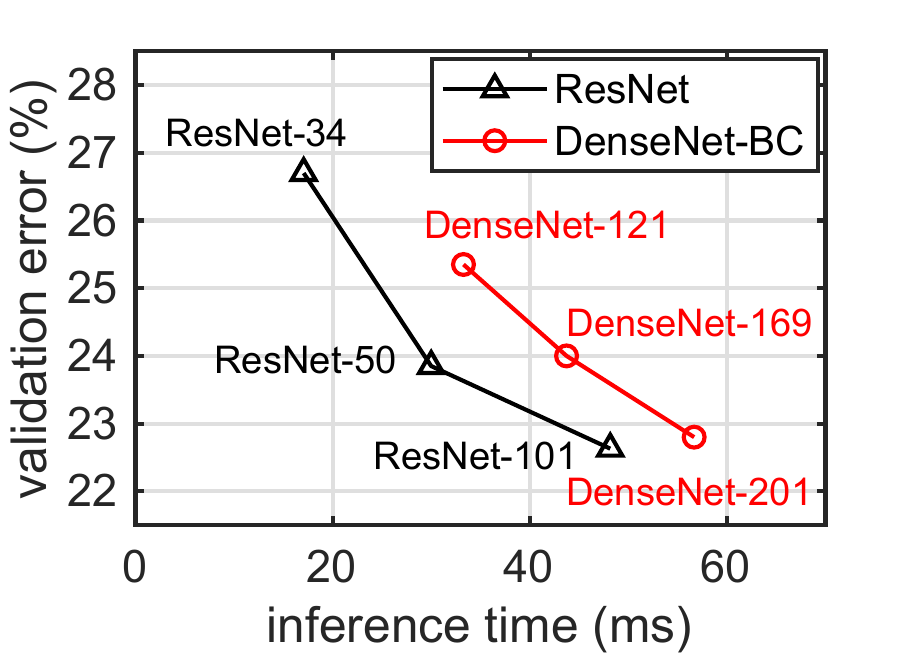}}
	\caption{ResNet and DenseNet Top-1 validation errors for different numbers of multiplications (\textit{left}) and inference times (\textit{right}). 
	  To measure the inference time, single NVIDIA Titan X (Pascal) is used and batch size is set to $16$. 
	  DenseNet has much fewer multiplications than ResNet, but its inference time is much longer.}
	\label{fig:intro_comparison}
\end{figure}

However, the bottleneck and dense block actually increase inference time even though the number of multiplications is reduced. 
\figurename~\ref{fig:intro_comparison} shows the number of multiplications and actual inference time for three models of ResNet and DenseNet. 
ResNet-50 has a number of multiplications similar to that of ResNet-34 thanks to the $1\times1$ convolution, but it takes $1.8\times$ longer than ResNet-34. 
This is because the bottleneck block of ResNet has four times larger activation map compared with basic residual block, so it causes four times more activation load from off-chip memory. 
Although DenseNet has a much smaller number of parameters and multiplications compared with ResNet, its inference time is much longer than that of ResNet. 
DenseNet has much smaller total activations than ResNet, but actual activation load of DenseNet is much larger than that of ResNet because the layers in DenseNet use output activations of all previous layers and thus actual activation load is much larger than the total activation size.

In this paper, we focus on the inference time reduction rather than parameter and multiplication reduction. 
To reduce the inference time, we propose the \textit{network recasting} method by transforming the network architecture for smaller activation load. 
We transform the network architecture through block-wise recasting of source blocks into target blocks. 
The recasting is done by training the target block to mimic the output activation of the source block, so the accuracy can be preserved after recasting. 
We can obtain a \textit{mixed-architecture network} by recasting parts of the trained network. 
By the mixed-architecture network, we mean a network having multiple types of block that can exploit the advantages of individual block types within a single network. 
In addition, we can use the network recasting method for network compression by recasting each block to a smaller one of the same type. 
We have achieved up to $3.2\times$ actual speedup with $0.22$\% top-5 accuracy loss on ILSVRC2012 dataset by the DenseNet-121 recasting.

\section{Related Works}

\paragraph{Network pruning}
To reduce the size and inference time of a trained network, several pruning methods such as weight pruning and filter pruning have been proposed. 
\citet{han2015learning} propose an iterative weight pruning method that removes connections and neurons according to the absolute value of parameters. 
\citet{guo2016dynamic} also propose iterative weight pruning that also gives a chance to restore connections for pruned weight. 
However, weight pruning methods generate sparse parameter matrices rather than smaller matrices, so its actual speedup is much less than the parameter reduction in general purpose hardware \cite{liu2015sparse}. 
The filter pruning methods reduce the size of parameter and activation matrices after the pruning, so they are more effective to accelerate the inference in any kinds of hardware. 
To find filters to be pruned, average percentage of zeros (APoZ), sum of absolute values, and reconstruction error of activation are used \cite{hu2016network,li2016pruning,luo2017thinet,he2017channel}. 
\citet{luo2017thinet} find and remove filters that have the smallest influence on the output activation of the next layer, and \citet{he2017channel} train a channel pruning mask minimizing the reconstruction error of current output activation and prune the channel of filters according to the trained mask.
\citet{liu2017learning} and \citet{luo2018autopruner} use channel scaling factor by adapting additional trainable parameters or squeeze-and-excitation layer \cite{hu2017squeeze}, and then prune filters according to the scaling factor.
Recently, \citet{lin2017runtime} use deep reinforcement learning to select pruning candidates at runtime. 

\paragraph{Knowledge distillation}
To train a smaller network with higher accuracy, mimic learning and knowledge distillation (KD) are introduced by \citet{ba2014deep} and \citet{hinton2015distilling}, respectively. 
These methods train a smaller network called \textit{student network} using logits of a large network called \textit{teacher network}. 
\citet{ba2014deep} train a student network by minimizing L2 loss between logits of student and teacher networks. 
\citet{hinton2015distilling} use logits of the teacher network to generate soft target, and train student network by minimizing cross-entropy loss with the soft target. 
It is hard to train a deep student network due to the vanishing-gradient problem, so several KD methods have been proposed to train a deep student network \cite{romero2014fitnets,luo2016face}. 
To train a thinner and deeper student network using KD, \citet{romero2014fitnets} propose hint training that trains a hidden layer with a convolutional regressor. 
\citet{luo2016face} make additional paths from a hidden layer to the output layer for gradient propagation without vanishing. 
In addition, \citet{zagoruyko2016paying} introduce the attention transfer method to reduce the number of residual blocks while conserving the accuracy. 
\citet{yim2017gift} also propose the residual block reduction method using the relationship between input and output activations. 
Also, there is a recent research to train the ResNet using logits of DenseNet \cite{furlanello2017born}.

\paragraph{Key differences}
Our work is for general recasting of neural networks. 
It can be used in various ways such as network size reduction or network type transformation. 
Compared to the previous work on network size reduction using weight/filter pruning, our work is different in that the inference process of the reduced network can be made significantly faster through the reduction of activations.
We also reveal the fact that reducing activation size is more important for inference speed than reducing the number of parameters.
Compared to other approaches using the knowledge distillation technique, our work is different in that the technique is applied sequentially to further enhance the accuracy.

\section{Network Recasting}
The network recasting method recasts a pre-trained network into a network of different type and/or size. 
Given the pre-trained teacher network, we transform each block (\textit{source block}) in the teacher network into a new block (\textit{target block}) of pre-defined type and size in the student network. 
The transformation is done by training the target block to generate output activations similar to those of the source block.
We call this process \textit{block recasting}. 
In this process, the source block can be considered as an unknown function, and the target block can be considered as a functional approximator similar to a multilayer perceptron \cite{hornik1991approximation}. 
After recasting all candidate blocks, we obtain the student network, which is faster than the teacher network while preserving the functionality or accuracy. 
We call the entire process \textit{network recasting}.

\subsection{Recasting from DenseNet to ResNet and ConvNet}
\begin{figure}[t]
\centering
	\includegraphics[width=0.85\columnwidth]{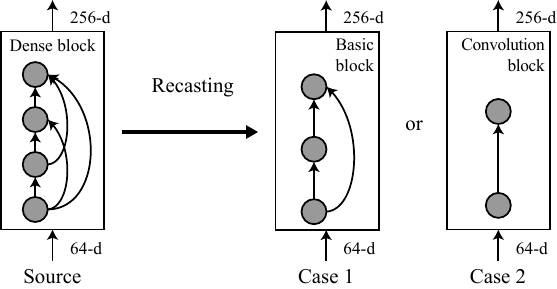}
	\caption{Block recasting of a dense block into a basic block (Case 1) and a convolution block (Case 2). 
	  The basic block has shorter inference time than the dense block because it has much smaller activation load. 
	  The convolution block is even faster than the basic block, but its capacity is much smaller and so it can cause accuracy loss.}
	\label{fig:dense}

\end{figure}
 
The DenseNet has a lot of activation load due to the dense connection, and by recasting the dense block into a basic residual block (we call the basic residual block as \textit{basic block} for simplicity), we can reduce the inference time. 
We consider a basic block consisting of two $3\times3$ convolution and shortcut as shown in \figurename~\ref{fig:dense}. 
Even though the basic block has more parameters and multiplications than the dense block, its activation load is much smaller and thus it is much faster. 
For more inference time reduction, we can recast the dense block into a single convolution block, although it can cause more accuracy loss because it has a very small capacity. 
\figurename~\ref{fig:dense} shows the two examples of recasting the first dense block in DenseNet-121.

\subsection{Recasting from ResNet to ConvNet}

\begin{figure}[t]
	\centering
	\includegraphics[width=0.85\columnwidth]{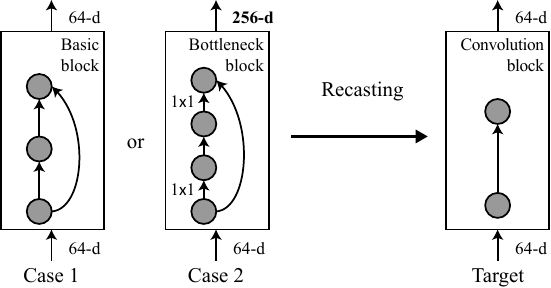}
	\caption{Block recasting of a residual block---basic block (Case 1) and bottleneck (Case 2)--- into a convolution block.
	  The recasting of the basic block keeps the same number of input and output channels. 
  However, since the bottleneck block uses a smaller number of channels for the feature extraction, we recast it into a convolution block that has the same number of input and output channels as the original $3\times3$ convolution.}
	\label{fig:res}
\end{figure}

\figurename~\ref{fig:res} illustrates the block recasting of a residual block into a convolution block. 
In the basic block, local features are extracted from the input activations using $3\times3$ filters, and thus, we recast the basic block into a $3\times3$ convolution block. 
Since the new convolution block has the same number of filters as the original basic block, the dimension of the output activations is not changed. 
However, in bottleneck block recasting, the dimension of the output activation is reduced as shown in \figurename~\ref{fig:res} (Case 2) for the first bottleneck block of ResNet-50.
Although the output activation becomes smaller, the number of linearly independent features is not changed because the second $1\times1$ convolution in the source block just combines its input activations linearly to extend the dimension of output activation.
Therefore, the next block in the student network still can reconstruct similar activation map.

\subsection{Compression} \label{sec:comp}

The network recasting can be used to compress the large network while preserving accuracy.
In this case, we assume that the network has redundancy such as ineffectual filters and redundant filters.
An \textit{ineffectual filter} denotes a filter that cannot extract any meaningful feature, and a \textit{redundant filter} denotes a filter that extracts a feature very similar to the one extracted by some other filter or a feature that can be obtained by combining features from other filters.
To remove those filters, previous approaches use APoZ \cite{hu2016network}, sum of absolute values of a filter \cite{li2016pruning}, or influence on next activations \cite{luo2017thinet,he2017channel} as the criteria, but redundant filters cannot be founded with those approaches. 
A possible approach is to find such redundant filters by checking the similarity between every pair of filters. 
However, it requires a huge amount of computations for similarity check and does not guarantee a good result. 
Instead, we recast a given source block into a smaller target block that has the same type as the source block.
Then we train the target block and the next block to reconstruct the output activation of the next block with smaller number of filters.
If the next block can reconstruct a similar output activation, the new target block can extract effective features for reconstruction.
For example, a convolution block is recast into another convolution block that has a smaller number of filters.
Then we train both the new convolution block and the next block to reconstruct the orginal activation map of the source next block.
After training, we can obtain a more effective filter set without any similarity or effectiveness check criteria.

\begin{table}[t]
\caption{Candidates for the network recasting.}
\centering
\begin{small}
	\begin{tabular}{@{ }c@{ }@{ }c@{ }@{ }c@{ }@{ }c@{ }}
\hline
Recasting Type & Source & Target & Dimension \\
\hline

Transformation & 
$\begin{array}{c} \text{Dense} \\ \text{Dense} \\ \text{Basic} \\ \text{Bottleneck} \end{array}$ &
$\begin{array}{c} \text{Basic} \\ \text{Convolution} \\ \text{Convolution} \\ \text{Convolution} \end{array}$ &
$\begin{array}{c} \text{Preserved} \\ \text{Preserved} \\ \text{Preserved} \\ \text{Reduced} \end{array}$ \\
\hline
Compression & 
$\begin{array}{c} \text{Basic} \\ \text{Convolution}  \end{array}$ &
$\begin{array}{c} \text{Basic} \\ \text{Convolution}  \end{array}$ &
$\begin{array}{c} \text{Reduced} \\ \text{Reduced}  \end{array}$ \\

\hline
\end{tabular}
\end{small}
\label{tab:recasting}
\end{table}

\subsection{Block Training}

\begin{figure}[t]
	\centering
	\includegraphics[width=\columnwidth]{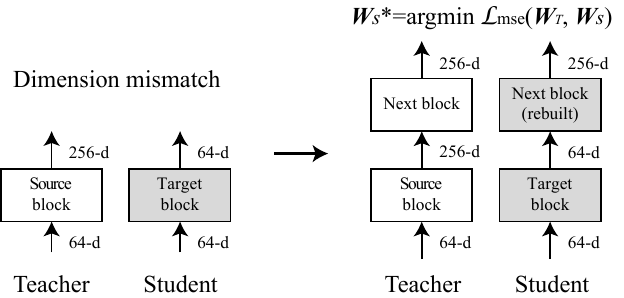}
	\caption{The dimension mismatch happens when the source block is recast into a smaller target block.
		The next block is used to match the dimension of output activation.
		After rebuilding the next block, both blocks are trained by minimizing $\mathcal{L}_{mse}(W_T, W_S)$.}
	\label{fig:mismatch}
\end{figure}

\begin{figure*}[t]
	\centering
	\includegraphics[width=\textwidth]{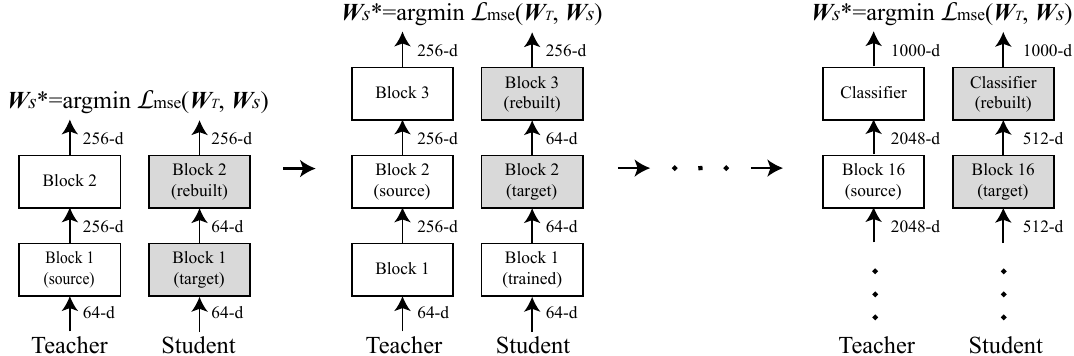}
	\caption{An example of sequential recasting for ResNet-50. 
	All blocks are recast in this example. 
	In each step, the target block and the next block (shaded blocks) are initialized randomly and trained by minimizing $\mathcal{L}_{mse}(W_T, W_S)$.}
	\label{fig:training}
\end{figure*}

For the target block to work properly, it should be trained with the source block as the teacher. 
We can easily train the target block by approximating the output activations to those of the source block if both blocks have the same dimension of output activations. 
However, dimension mismatch happens in the block recasting especially when we reduce the number of channels for network size reduction.
Table~\ref{tab:recasting} shows the recasting cases that we handle in this paper.
To avoid the dimension mismatch problem, when training a target block, we train the target block together with the next block by approximating the output activations of the next block as shown in \figurename~\ref{fig:mismatch}. 
The next block is rebuilt from the corresponding source block by reducing the filter size when the target block has a smaller number of channels. 
Both the target block and the next block are initialized randomly and trained to minimize the loss of mean-square error (MSE) between teacher's and student's activations given by,
\begin{equation}
	\label{eq:mseloss}
	  \mathcal{L}_{mse}(W_T, W_S) = 
	  \frac{1}{N}{\left\|{A(x;W_T) - A(x;W_S)}\right\|}_2^2,
\end{equation}
where $A$ means the activation of the next block, and $x$ is the input data. 
$W_T$ and $W_S$ indicate parameters of teacher network and student network, respectively. 
$N$ denotes the size of an output activation of the next block. 

\subsection{Sequential Recasting and Fine-tuning} \label{sec:process}

To recast the entire network, we apply the block recasting method sequentially. 
\figurename~\ref{fig:training} shows an example of sequential recasting method. 
The type and dimension of the first (target) block of the student network are determined, and then the second block is rebuilt from the second block of the teacher network; if there is no dimension mismatch, the second block will be the same as that of the teacher network. 
The two blocks are initialized randomly and trained by minimizing $\mathcal{L}_{mse}(W_T, W_S)$. 
Now, the second block becomes the target. 
Thus, its type and dimension are determined, the third block is rebuilt, and both blocks are initialized randomly. 
To train the second and third blocks, we reuse the trained first block. 
The first block is already trained in the previous step, but it still has approximation errors. 
We can reduce the effect of its errors by training both the previous and current blocks. 
Therefore, three blocks are trained in the second step by minimizing $\mathcal{L}_{mse}(W_T, W_S)$. 
This process is continued for the following blocks until the last block is recast as a new block. 
We can select arbitrary blocks as candidates for recasting so that the student network can consist of multiple types of block. 
For example, the student network can have both residual and dense blocks when only the first dense block is recast into a residual block. 
We call the network that has multiple types of block as \textit{mixed-architecture network}. 
The mixed-architecture network can have advantages of both blocks. 
For example, by mixing dense blocks and residual blocks, we can obtain a mixed-architecture network that is faster than DenseNet and has fewer parameters than ResNet.

The block-by-block sequential recasting has two advantages. 
First, the functionality of each block is much simpler than that of the whole network. 
Thus, it is easier to approximate the functionality of each block. 
By approximating each of easier sub-functions, we can finally obtain the student network with smaller approximation error. 
Secondly, sequential recasting can alleviate the vanishing-gradient problem. 
When the source block is recast as a convolution block, the student network cannot be trained well due to the gradient vanishing. 
However, sequential recasting has very short gradient paths from the output activation to the target block, so it can be trained well. 
Therefore, we can obtain the student network with higher accuracy using sequential recasting.

\begin{table*}[t]
\caption{Error rates (\%) of architecture transform results on CIFAR datasets. (B/M: billion/million)}
\centering
\begin{small}
\begin{tabular}{lcrrrrrr}
\hline
\multicolumn{1}{c}{Method} & \multicolumn{1}{c}{Type} & \multicolumn{1}{c}{C10+} & \multicolumn{1}{c}{C100+} & \multicolumn{1}{c}{Params} & \multicolumn{1}{c}{Mults} & \multicolumn{1}{c}{Acts/image} & \multicolumn{1}{c}{Time/image} \\
\hline
\hline
\multicolumn{8}{c}{ResNet-56} \\
\hline
Baseline & & $7.02$ & $30.89$ & $0.85$M ($1.0\times$) & $125.75$M ($1.0\times$) & $0.56$M ($1.0\times$) & $1.05$ms \\[1ex]
Recasting & Conv & $\textbf{6.75}$ & $\textbf{32.14}$ & $0.41$M ($2.1\times$) & $61.78$M ($2.0\times$) & $0.27$M ($2.0\times$) & $0.50$ms \\
KD & Conv & $9.43$ & $33.22$ & $0.41$M ($2.1\times$) & $61.78$M ($2.0\times$) & $0.27$M ($2.0\times$) & $0.50$ms \\
Backprop & Conv & $10.61$ & $37.85$ & $0.41$M ($2.1\times$) & $61.78$M ($2.0\times$) & $0.27$M ($2.0\times$) & $0.50$ms \\
\hline
\multicolumn{8}{c}{ResNet-83} \\
\hline
Baseline & & $6.34$ & $28.13$ & $0.83$M ($1.0\times$) & $125.09$M ($1.0\times$) & $1.69$M ($1.0\times$) & $1.51$ms \\[1ex]
Recasting & Conv & $\textbf{6.90}$ & $\textbf{31.04}$ & $0.41$M ($2.0\times$) & $61.78$M ($2.0\times$) & $0.27$M ($6.2\times$) & $0.95$ms \\
KD & Conv & $8.95$ & $32.75$ & $0.41$M ($2.0\times$) & $61.78$M ($2.0\times$) & $0.27$M ($6.2\times$) & $0.95$ms \\
Backprop & Conv & $9.77$ & $37.14$ & $0.41$M ($2.0\times$) & $61.78$M ($2.0\times$) & $0.27$M ($6.2\times$) & $0.95$ms \\
\hline
\multicolumn{8}{c}{WRN-28-10} \\
\hline
Baseline & & $4.06$ & $19.54$ & $36.45$M ($1.0\times$) & $5.24$B ($1.0\times$) & $2.52$M ($1.0\times$) & $0.81$ms \\[1ex]
Recasting & Conv & $\textbf{4.11}$ & $\textbf{19.74}$ & $4.86$M ($7.5\times$) & $1.17$B ($4.5\times$) & $0.90$M ($2.8\times$) & $0.41$ms \\
KD & Conv & $4.40$ & $19.94$ & $4.86$M ($7.5\times$) & $1.17$B ($4.5\times$) & $0.90$M ($2.8\times$) & $0.41$ms \\
Backprop & Conv & $4.67$ & $20.90$ & $4.86$M ($7.5\times$) & $1.17$B ($4.5\times$) & $0.90$M ($2.8\times$) & $0.41$ms \\
\hline
\multicolumn{8}{c}{DenseNet-100} \\
\hline
Baseline & & $5.11$ & $23.62$ & $0.74$M ($1.0\times$) & $0.29$B ($1.0\times$) & $4.41$M ($1.0\times$) & $2.12$ms \\[1ex]
Recasting & Basic & $4.91$ & $\textbf{22.39}$ & $2.53$M ($0.3\times$) & $0.77$B ($0.4\times$) & $0.89$M ($4.9\times$) & $0.27$ms \\
KD & Basic & $\textbf{4.71}$ & $22.71$ & $2.53$M ($0.3\times$) & $0.77$B ($0.4\times$) & $0.89$M ($4.9\times$) & $0.27$ms \\
Backprop & Basic & $5.39$ & $24.57$ & $2.53$M ($0.3\times$) & $0.77$B ($0.4\times$) & $0.89$M ($4.9\times$) & $0.27$ms \\[1ex]
Recasting & Conv & $6.82$ & $\textbf{25.60}$ & $0.87$M ($0.9\times$) & $0.19$B ($1.5\times$) & $0.51$M ($8.6\times$) & $0.16$ms \\
KD & Conv & $\textbf{6.75}$ & $26.52$ & $0.87$M ($0.9\times$) & $0.19$B ($1.5\times$) & $0.51$M ($8.6\times$) & $0.16$ms \\
Backprop & Conv & $8.11$ & $30.05$ & $0.87$M ($0.9\times$) & $0.19$B ($1.5\times$) & $0.51$M ($8.6\times$) & $0.16$ms \\
\hline
\end{tabular}
\end{small}
\label{tab:cifar_transform}
\end{table*}

\begin{figure}[t]
\centering
	\subfloat{{\includegraphics[width=0.4\columnwidth]{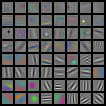}\label{fig:filter_teacher}}}
	\qquad
	\subfloat{{\includegraphics[width=0.2528\columnwidth]{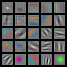}\label{fig:filter_student}}}
	\caption{Visualization of filters in the first layer of AlexNet (\textit{left}) and a student network (\textit{right}). 
	  	      Redundant filters are removed after network recasting.}
	\label{fig:filter_reduction}
\end{figure}

After finishing sequential recasting, we use the knowledge distillation approach to fine-tune the student network. % for the main classification task. 
There are approximation errors after sequential recasting, and we can reduce the effect of those errors by training the whole network. 
We train the student network with logits of the teacher network and ground truth. 
Thus, our knowledge distillation (KD) loss is defined by
\begin{equation}
	\label{eq:KDloss}
	\mathcal{L}_{kd}(W_T, W_S) = \mathcal{L}_{mse\_logit}(W_T, W_S) + \mathcal{L}_{ce}(y_{true}, W_S),
\end{equation}
where $\mathcal{L}_{mse\_logit}$ is the MSE loss for the logits, and $\mathcal{L}_{ce}$ is the cross-entropy loss between the given label $y_{true}$ and softmax output of the student network that is parameterized by $W_s$.

\section{Experiments}

\begin{table*}[t]
\caption{Error rates (\%) of compression results on CIFAR datasets. (B/M: billion/million)}
\centering
\begin{small}
\begin{tabular}{lcrrrrrr}
\hline
\multicolumn{1}{c}{Method} & \multicolumn{1}{c}{Type} & \multicolumn{1}{c}{C10+} & \multicolumn{1}{c}{C100+} & \multicolumn{1}{c}{Params} & \multicolumn{1}{c}{Mults} & \multicolumn{1}{c}{Acts/image} & \multicolumn{1}{c}{Time/image} \\
\hline
\hline
\multicolumn{8}{c}{VGG-16} \\
\hline
Baseline & & $6.85$ & $28.80$ & $14.71$M ($1.0\times$) & $313.20$M ($1.0\times$) & $0.31$M ($1.0\times$) & $0.37$ms \\
[1ex]
Recasting & Conv & $\textbf{8.31}$ & $\textbf{31.56}$ & $2.36$M ($6.2\times$) & $50.63$M ($6.2\times$) & $0.13$M ($2.4\times$) & $0.31$ms \\
KD & Conv & $9.24$ & $33.14$ & $2.36$M ($6.2\times$) & $50.63$M ($6.2\times$) & $0.13$M ($2.4\times$) & $0.31$ms \\
Backprop & Conv & $8.71$ & $35.13$ & $2.36$M ($6.2\times$) & $50.63$M ($6.2\times$) & $0.13$M ($2.4\times$) & $0.31$ms \\
\hline
\multicolumn{8}{c}{WRN-28-10} \\
\hline
Baseline & & $4.06$ & $19.54$ & $36.45$M ($1.0\times$) & $5.24$B ($1.0\times$) & $2.52$M ($1.0\times$) & $0.81$ms \\
[1ex]
Recasting & Basic & $\textbf{5.18}$ & $\textbf{24.13}$ & $1.46$M ($24.9\times$) & $0.21$B ($24.5\times$) & $0.52$M ($4.9\times$) & $0.56$ms \\
KD & Basic & $5.48$ & $25.28$ & $1.46$M ($24.9\times$) & $0.21$B ($24.5\times$) & $0.52$M ($4.9\times$) & $0.56$ms \\
Backprop & Basic & $5.39$ & $25.78$ & $1.46$M ($24.9\times$) & $0.21$B ($24.5\times$) & $0.52$M ($4.9\times$) & $0.56$ms \\ 
\hline

\end{tabular}
\end{small}
\label{tab:cifar_compress}
\end{table*}

We conducted several experiments for the network recasting. 
For the experiments, we used CIFAR and ILSVRC2012 dataset and four kinds of network architectures; ResNet \cite{he2016deep}, Wide ResNet (WRN) \cite{zagoruyko2016wide}, DenseNet \cite{huang2017densely}, and VGG-16 \cite{simonyan2014very}. 
We adopted batch normalization \cite{ioffe2015batch} for all networks, because it was also effective for block-wise training. 
The network recasting was implemented on the \textit{PyTorch} framework. 
We used the Xavier initializer \cite{glorot2010understanding} in all experiments. 
We used SGD with Nesterov momentum \cite{sutskever2013importance} to train the teacher network and used Adam optimizer \cite{kingma2014adam} for the network recasting. 
In addition, we trained the student network with KD and back propagation from scratch using SGD with Nesterov momentum for the comparison.

\subsection{Visualization of Filter Reduction}
 
The network recasting can be used for network compression; it can remove redundant filters as well as ineffectual filters.
To show the filter reduction, we compressed only the first layer of AlexNet and visualized the filter set in \figurename~\ref{fig:filter_reduction}. 
The first layer of the original AlexNet had $64$ filters, but we decreased the number to $25$ in the student network. 
Then we trained the first block of the student network for eight epochs, and fine-tuned the entire student network; the learning rates for the recasting of the first block and the fine-tuning were $0.0005$ and $0.0001$, respectively.
Every five epochs, the running rates were divided by 10.
\figurename~\ref{fig:filter_reduction} shows filters extracted from the first layers of the teacher and student networks. 
Filters of the teacher network consist of many ineffectual and redundant filters, but those are eliminated as shown in \figurename~\ref{fig:filter_reduction}. 
In addition, the student network achieves the top-1 error of $44.20$\% and the top-5 error of $21.54$\%. The top-1 and the top-5 errors increase by only $0.72$\% and $0.61$\%, respectively. 
Note that it is hard to remove many filters without accuracy loss because AlexNet has a relatively large ($11\times11$) filters. 
The filter size is related to the dimension of filter vector, and many more filters are required to span the vector space as the filter size increases. 
As expected, we could remove many more filters on both VGG-16 and ResNet, which have only $3\times3$ filters. 

\subsection{CIFAR}

For CIFAR dataset, we used ResNet-56, ResNet-83, WRN-28-10, DenseNet-100, and VGG-16.
Especially, ResNet-83 has the same number of blocks with ResNet-56, but consists of bottleneck blocks. 
In addition, we used a modified version of VGG16, which has only one hidden fully-connected layer with $512$ neurons. 
Teacher networks were trained from scratch using back propagation. 
We used CIFAR-10 and 100 dataset with the standard data augmentation, which consists of four pixel zero-padding and random cropping, and horizontal flipping with $0.5$ probability.

In CIFAR experiments, we recast all blocks of teacher networks, so there is no mixed-architecture result. 
We counted the number of parameters, multiplications, and activation loads for the convolution operation. 
Especially, we reported the activation load of a single image in \tablename~\ref{tab:cifar_transform}. 
\tablename~\ref{tab:cifar_transform} shows the architecture transformation results. 
The network recasting achieved similar accuracy with the teacher network, and activation access is reduced significantly. 
It shows lower test error compared to other methods in all network architectures. 
When networks were recast into a plain convolutional network, the network recasting achieved much lower test error compared with both KD and back propagation. 
The sequential recasting can alleviate the vanishing-gradient problem, so its results outperformed the others. 

We also compressed the VGG-16 and WRN-28-10 using the network recasting. 
In this experiment, source blocks were recast into $2.5\times$ and $5\times$ smaller blocks in VGG-16 and WRN-28-10, respectively. 
\tablename~\ref{tab:cifar_compress} shows compression results of both networks. 
The network recasting achieved the smallest accuracy loss compared with other methods. 
Especially, network recasting achieved $1.58$\% and $3.57$\% lower test error compared with KD and back propagation in VGG-16 compression on CIFAR-100.

\begin{table*}[t]
\caption{Error rate (\%) of network recasting results on ILSVRC2012. (B/M: billion/million)}
\centering
\begin{small}
\begin{tabular}{lrrrrrrr}
\hline
\multicolumn{1}{c}{Method} & \multicolumn{1}{c}{Top1} & \multicolumn{1}{c}{Top5} & \multicolumn{1}{c}{Params} & \multicolumn{1}{c}{Mults} & \multicolumn{1}{c}{Acts/image} & \multicolumn{1}{c}{Time/image} & \multicolumn{1}{c}{Time/bacth}\\
\hline
\hline
\multicolumn{8}{c}{ResNet-50} \\
\hline
Baseline & $23.85$ & $7.13$ & $25.50$M & $4.09$B & $11.57$M & $6.16$ms & $107.17$ms \\[1ex]
Recasting(C) & $30.74$ & $10.39$ & $10.29$M & $1.71$B & $2.53$M & $\textbf{2.12}$\textbf{ms} & $\textbf{37.21}$\textbf{ms} \\
Recasting(C+R$_{bt}$) & $\textbf{25.00}$ & $\textbf{7.71}$ & $21.72$M & $2.40$B & $3.69$M & $3.79$ms & $49.97$ms \\
KD(C+R$_{bt}$) & $27.00$ & $8.30$ & $21.72$M & $2.40$B & $3.69$M & $3.79$ms & $49.97$ms \\
\hline
\multicolumn{8}{c}{DenseNet-121} \\
\hline
Baseline & $25.57$ & $8.03$ & $7.89$M & $2.75$B & $16.52$M & $12.73$ms & $111.31$ms \\[1ex]
Recasting(R$_{bs}$) & $26.42$ & $8.25$ & $32.23$M & $8.15$B & $5.32$M & $\textbf{3.95}$\textbf{ms} & $\textbf{81.17}$\textbf{ms} \\
Recasting(R$_{bs}$+D) & $\textbf{24.87}$ & $\textbf{7.59}$ & $10.42$M & $5.72$B & $9.15$M & $9.40$ms & $88.94$ms \\
KD(R$_{bs}$+D) & $24.90$ & $7.65$ & $10.42$M & $5.72$B & $9.15$M & $9.40$ms & $88.94$ms \\
\hline
\hline
\multicolumn{8}{c}{VGG-16} \\
\hline
Baseline & $26.63$ & $8.50$ & $138.34$M & $15.47$B & $15.09$M & $6.17$ms & $200.47$ms \\[1ex]
Recasting(C\_P) & $\textbf{28.25}$ & $\textbf{9.41}$ & $81.93$M & $4.73$B & $8.27$M & $\textbf{3.45}$\textbf{ms} & $116.45$ms \\
Recasting(C\_A) & $30.05$ & $10.38$ & $120.61$M & $3.12$B & $3.30$M & $3.61$ms & $\textbf{63.52}$\textbf{ms} \\
\hline
\end{tabular}
\end{small}
\label{tab:ilsvrc}
\end{table*}

\begin{table*}[t]
\caption{Comparison of error rate (\%) with previous works on ILSVRC2012. (B/M: billion/million)}
\centering
\begin{small}
\begin{tabular}{lcccccc}
\hline
\multicolumn{1}{c}{Method} & \multicolumn{1}{c}{Top1} & \multicolumn{1}{c}{Top5} & \multicolumn{1}{c}{Params} & \multicolumn{1}{c}{Mults} & \multicolumn{1}{c}{Acts/batch} & \multicolumn{1}{c}{Actual speed-up}\\
\hline
\hline
\multicolumn{7}{c}{ResNet-50} \\
\hline
Recasting(C+R$_{bt}$) & $\textbf{25.00}$ & $\textbf{7.71}$ & $21.72$M & $2.40$B & $236.16$M & $\textbf{2.1}\times$ \\
ThiNet-30 \cite{luo2017thinet} & $31.58$ & $11.7$ & $8.66$M & $1.10$B & - & $1.3\times$  \\
AutoPruner ($r=0.3$) \cite{luo2018autopruner} & $27.47$ & $8.89$ & - & $1.32$B & - & - \\
\hline
\hline
\multicolumn{7}{c}{VGG-16} \\
\hline
Recasting(C\_A) & $\textbf{30.05}$ & $\textbf{10.38}$ & $120.61$M & $3.12$B & $220.61$M & $\textbf{3.2}\times$ \\
ThiNet-Conv \cite{luo2017thinet} & $30.20$ & $10.47$ & $131.44$M & $4.79$B & - & $2.5\times$  \\
RNP ($3\times$) \cite{lin2017runtime} & - & $12.42$ & - & - & - & $2.3\times$ \\
Channel Pruning ($3\times$) \cite{he2017channel} &- & $11.10$ & - & - & - & $2.5\times$ \\
AutoPruner ($r=0.4$) \cite{luo2018autopruner} & $31.57$ & $11.57$ & - & $4.09$B & - & - \\
\hline
\end{tabular}
\end{small}
\label{tab:comparison}
\end{table*}

The born again network (BAN) proposed by \cite{furlanello2017born} also trains ResNet student using logits of DenseNet teacher. 
However, they proposed only switching DenseNet with ResNet, and the test error of BAN will be higher than that of network recasting because BAN only uses the KD method as shown in \tablename~\ref{tab:cifar_transform} and \ref{tab:cifar_compress}. 
We propose any to any architecture transformation, and deep student networks that have only convolution blocks can also be trained well by applying sequential recasting because it can alleviate the vanishing-gradient problem. 
In addition, we also propose mixed-architecture network, which can also be trained well by using the proposed network recasting.

\subsection{ILSVRC2012}

For ILSVRC2012 dataset, we used the pre-trained ResNet-50, DenseNet-121, and VGG-16 available from \textit{torchvision} which is one of the \textit{PyTorch} packages. 
These pre-trained networks were used as the teacher networks. 
We recast the blocks of ResNet-50 into convolution blocks, and the blocks of DenseNet-121 into basic blocks. 
In addition, we recast only parts of these networks to obtain mixed-architecture networks. 
In \tablename~\ref{tab:ilsvrc}, \textit{Recasting(C)} indicates that the student network only has convolution blocks, and \textit{Recasting(C+R$_{bt}$)} denotes that the student network has both convolution and bottleneck blocks. 
In the same way, \textit{Recasting(R$_{bs}$)} and \textit{Recasting(R$_{bs}$+D)} denotes that the student networks consist of only basic blocks and both basic blocks and dense blocks, respectively.
KD(C+R$_{bt}$) and KD(R$_{bs}$+D) have the same network architecture as Recasting(C+R$_{bt}$) and Recasting(R$_{bs}$+D) respectively, but those are trained with only KD method.
For the VGG-16 compression, we used two criteria: higher parameter reduction (\textit{Recasting(C\_P)}) and higher activation reduction (\textit{Recasting(C\_A)}). 
In addition, we measured the actual inference time for all networks on an NVIDIA Titan X (Pascal) GPU, and batch sizes were set to 1 and 64.

We measured the training time for Recasiting(C+R$_{bt}$), KD(C+R$_{bt}$), Recasting(R$_{bs}$+D) and KD(R$_{bs}$+D) to compare the training time and accuracy.
Recasting(C+R$_{bt}$) took $7.6$ days, and KD(C+R$_{bt}$) took $6.3$ days on a GPU.
Compared to KD(C+R$_{bt}$), Recasting(C+R$_{bt}$) took $20$\% longer, but achieved $2.00$\%p and $0.59$\%p improvement in top-1 and top-5 accuracy, respectively.
On the other hand, Recasting(R$_{bs}$+D) took $3.9$ days, while KD(R$_{bs}$+D) took $8.9$ days with similar accuracy.
Those results show that network recasting can achieve higher accuracy with slightly longer training time for a deep network and shorter training time with similar accuracy for a shallow network.

As shown in \tablename~\ref{tab:ilsvrc}, the network recasting significantly reduced the inference time in all experiments. 
Recasting(C) and Recasting(R$_{bs}$) achieved $2.9\times$ and $3.2\times$ inference time reduction for a single image compared with original ResNet-50 and DenseNet- 121, respectively. 
Moreover, mixed-architecture networks also achieved significant inference time reduction with smaller accuracy loss. 
For the batch processing, Recasting(C+R$_{bt}$) achieved $2.1\times$ time reduction with $0.58$\% top-5 accuracy loss compared to Baseline, and Recasting(R$_{bs}$+D) achieved $1.3\times$ time reduction even with $0.44$\% higher top-5 accuracy. 
In particular, Recasting(R$_{bs}$+D) achieved similar accuracy and inference time with $3.1\times$ fewer parameters compared to Recasting(R$_{bs}$). 
In VGG-16 compression, Recasting(C\_P) and Recasting(C\_A) achieved $1.7\times$ parameter reduction and $4.6\times$ activation reduction with $0.91$\% and $2.05$\% top-5 accuracy loss, respectively. 
Recasting(C\_A) achieved $3.2\times$ inference time reduction compared to the baseline. 

We compared our results with several previous approaches \cite{luo2017thinet,he2017channel,lin2017runtime,luo2018autopruner}. 
For the comparison, we used batch inference time because previous approaches have reported inference time only for the batch processing. 
\tablename~\ref{tab:comparison} shows that the network recasting achieved much higher inference time reduction. 
In ResNet-50, Recasting(C+R$_{bt}$) achieved lower error rate and much higher actual speedup compared with ThiNet \cite{luo2017thinet}. 
ThiNet only reduced filters and multiplications in $3\times3$ convolution of bottleneck blocks, so it cannot accelerate the inference effectively because activation load is still large. 
However, the network recasting can reduce the activation load effectively, so it achieved $2.1\times$ actual speedup with smaller accuracy loss.
\citet{luo2018autopruner} does not mention actual-speedup, but we can guess that our network recasting result is much faster than their AutoPruner result because they cannot remove the $1\times1$ convolution. 
For the VGG-16 compression, the network recasting also achieves much higher speedup with lower error rate compared to previous approaches. 
It also achieves higher parameter and multiplication reduction with similar accuracy compared to others.

\section{Conclusion}
In this paper, we proposed network recasting as a universal method for network architecture transformation. 
This method can accelerate network inference by transforming the network (teacher) to a more efficient one (student). 
We could recast residual and dense blocks into convolution and residual blocks, respectively, to achieve much higher actual speedup at small accuracy loss. 
By recasting blocks sequentially, the student network can be trained well even though there is no shortcut or dense connection. 
In addition, our method can recast arbitrary blocks, thereby producing a mixed-architecture network. 
The mixed-architecture networks produced as such achieved $2.1\times$ inference time with $0.58$\% top-5 accuracy loss compared to original ResNet-50, and also achieved $1.3\times$ inference time reduction with $0.44$\% higher top-5 accuracy on DenseNet-121 recasting.
We also applied the network recasting for the purpose of compression and achieved higher compression ratio and speedup compared to previous approaches. 
Our method can be applied to various kinds of network architecture to transform it into various kinds of target network architecture.

\section{Acknowledgments}
This work was supported by Samsung Advanced Institute of Technology.

\bibliographystyle{aaai}
\bibliography{AAAI-YuJ.3900}

\begin{thebibliography}{}

\bibitem[\protect\citeauthoryear{Ba and Caruana}{2014}]{ba2014deep}
Ba, J., and Caruana, R.
\newblock 2014.
\newblock Do deep nets really need to be deep?
\newblock In {\em NIPS},  2654--2662.

\bibitem[\protect\citeauthoryear{Furlanello \bgroup et al\mbox.\egroup
  }{2018}]{furlanello2017born}
Furlanello, T.; Lipton, Z.~C.; Amazon, A.; Itti, L.; and Anandkumar, A.
\newblock 2018.
\newblock Born again neural networks.
\newblock In {\em ICML},  1607--1616.

\bibitem[\protect\citeauthoryear{Glorot and
  Bengio}{2010}]{glorot2010understanding}
Glorot, X., and Bengio, Y.
\newblock 2010.
\newblock Understanding the difficulty of training deep feedforward neural
  networks.
\newblock In {\em AISTATS},  249--256.

\bibitem[\protect\citeauthoryear{Goodfellow \bgroup et al\mbox.\egroup
  }{2016}]{goodfellow2016deep}
Goodfellow, I.; Bengio, Y.; Courville, A.; and Bengio, Y.
\newblock 2016.
\newblock {\em Deep learning}, volume~1.
\newblock MIT press Cambridge.

\bibitem[\protect\citeauthoryear{Guo, Yao, and Chen}{2016}]{guo2016dynamic}
Guo, Y.; Yao, A.; and Chen, Y.
\newblock 2016.
\newblock Dynamic network surgery for efficient {DNNs}.
\newblock In {\em NIPS},  1379--1387.

\bibitem[\protect\citeauthoryear{Han \bgroup et al\mbox.\egroup
  }{2015}]{han2015learning}
Han, S.; Pool, J.; Tran, J.; and Dally, W.
\newblock 2015.
\newblock Learning both weights and connections for efficient neural network.
\newblock In {\em NIPS},  1135--1143.

\bibitem[\protect\citeauthoryear{He \bgroup et al\mbox.\egroup
  }{2016}]{he2016deep}
He, K.; Zhang, X.; Ren, S.; and Sun, J.
\newblock 2016.
\newblock Deep residual learning for image recognition.
\newblock In {\em CVPR},  770--778.

\bibitem[\protect\citeauthoryear{He, Zhang, and Sun}{2017}]{he2017channel}
He, Y.; Zhang, X.; and Sun, J.
\newblock 2017.
\newblock Channel pruning for accelerating very deep neural networks.
\newblock In {\em ICCV},  1389--1397.

\bibitem[\protect\citeauthoryear{Hinton, Vinyals, and
  Dean}{2014}]{hinton2015distilling}
Hinton, G.; Vinyals, O.; and Dean, J.
\newblock 2014.
\newblock Distilling the knowledge in a neural network.
\newblock In {\em NIPS Deep Learning and Representation Learning Workshop}.

\bibitem[\protect\citeauthoryear{Hornik}{1991}]{hornik1991approximation}
Hornik, K.
\newblock 1991.
\newblock Approximation capabilities of multilayer feedforward networks.
\newblock {\em Neural networks} 4(2):251--257.

\bibitem[\protect\citeauthoryear{Hu \bgroup et al\mbox.\egroup
  }{2016}]{hu2016network}
Hu, H.; Peng, R.; Tai, Y.-W.; and Tang, C.-K.
\newblock 2016.
\newblock Network trimming: A data-driven neuron pruning approach towards
  efficient deep architectures.
\newblock {\em arXiv preprint arXiv: 1607.03250}.

\bibitem[\protect\citeauthoryear{Hu, Shen, and Sun}{2018}]{hu2017squeeze}
Hu, J.; Shen, L.; and Sun, G.
\newblock 2018.
\newblock Squeeze-and-excitation networks.
\newblock In {\em CVPR},  7132--7141.

\bibitem[\protect\citeauthoryear{Huang \bgroup et al\mbox.\egroup
  }{2017}]{huang2017densely}
Huang, G.; Liu, Z.; Weinberger, K.~Q.; and van~der Maaten, L.
\newblock 2017.
\newblock Densely connected convolutional networks.
\newblock In {\em CVPR},  4700--4708.

\bibitem[\protect\citeauthoryear{Ioffe and Szegedy}{2015}]{ioffe2015batch}
Ioffe, S., and Szegedy, C.
\newblock 2015.
\newblock Batch normalization: Accelerating deep network training by reducing
  internal covariate shift.
\newblock In {\em ICML},  448--456.

\bibitem[\protect\citeauthoryear{Kingma and Ba}{2015}]{kingma2014adam}
Kingma, D.~P., and Ba, J.
\newblock 2015.
\newblock {Adam}: A method for stochastic optimization.
\newblock In {\em ICLR}.

\bibitem[\protect\citeauthoryear{Larsson, Maire, and
  Shakhnarovich}{2017}]{larsson2016fractalnet}
Larsson, G.; Maire, M.; and Shakhnarovich, G.
\newblock 2017.
\newblock Fractalnet: Ultra-deep neural networks without residuals.
\newblock In {\em ICLR}.

\bibitem[\protect\citeauthoryear{Li \bgroup et al\mbox.\egroup
  }{2016}]{li2016pruning}
Li, H.; Kadav, A.; Durdanovic, I.; Samet, H.; and Graf, H.~P.
\newblock 2016.
\newblock Pruning filters for efficient convnets.
\newblock In {\em ICLR}.

\bibitem[\protect\citeauthoryear{Lin \bgroup et al\mbox.\egroup
  }{2017}]{lin2017runtime}
Lin, J.; Rao, Y.; Lu, J.; and Zhou, J.
\newblock 2017.
\newblock Runtime neural pruning.
\newblock In {\em NIPS},  2181--2191.

\bibitem[\protect\citeauthoryear{Lin, Chen, and Yan}{2014}]{lin2013network}
Lin, M.; Chen, Q.; and Yan, S.
\newblock 2014.
\newblock Network in network.
\newblock In {\em ICLR}.

\bibitem[\protect\citeauthoryear{Liu \bgroup et al\mbox.\egroup
  }{2015}]{liu2015sparse}
Liu, B.; Wang, M.; Foroosh, H.; Tappen, M.; and Pensky, M.
\newblock 2015.
\newblock Sparse convolutional neural networks.
\newblock In {\em CVPR},  806--814.

\bibitem[\protect\citeauthoryear{Liu \bgroup et al\mbox.\egroup
  }{2017}]{liu2017learning}
Liu, Z.; Li, J.; Shen, Z.; Huang, G.; Yan, S.; and Zhang, C.
\newblock 2017.
\newblock Learning efficient convolutional networks through network slimming.
\newblock In {\em ICCV},  2755--2763.

\bibitem[\protect\citeauthoryear{Luo and Wu}{2018}]{luo2018autopruner}
Luo, J.-H., and Wu, J.
\newblock 2018.
\newblock {AutoPruner}: An end-to-end trainable filter pruning method for
  efficient deep model inference.
\newblock {\em arXiv preprint arXiv: 1805.08941}.

\bibitem[\protect\citeauthoryear{Luo \bgroup et al\mbox.\egroup
  }{2016}]{luo2016face}
Luo, P.; Zhu, Z.; Liu, Z.; Wang, X.; and Tang, X.
\newblock 2016.
\newblock Face model compression by distilling knowledge from neurons.
\newblock In {\em AAAI},  3560--3566.

\bibitem[\protect\citeauthoryear{Luo, Wu, and Lin}{2017}]{luo2017thinet}
Luo, J.-H.; Wu, J.; and Lin, W.
\newblock 2017.
\newblock {ThiNet}: A filter level pruning method for deep neural network
  compression.
\newblock In {\em ICCV},  5058--5066.

\bibitem[\protect\citeauthoryear{Romero \bgroup et al\mbox.\egroup
  }{2015}]{romero2014fitnets}
Romero, A.; Ballas, N.; Kahou, S.~E.; Chassang, A.; Gatta, C.; and Bengio, Y.
\newblock 2015.
\newblock {FitNets}: Hints for thin deep nets.
\newblock In {\em ICLR}.

\bibitem[\protect\citeauthoryear{Simonyan and
  Zisserman}{2015}]{simonyan2014very}
Simonyan, K., and Zisserman, A.
\newblock 2015.
\newblock Very deep convolutional networks for large-scale image recognition.
\newblock In {\em ICLR}.

\bibitem[\protect\citeauthoryear{Sutskever \bgroup et al\mbox.\egroup
  }{2013}]{sutskever2013importance}
Sutskever, I.; Martens, J.; Dahl, G.; and Hinton, G.
\newblock 2013.
\newblock On the importance of initialization and momentum in deep learning.
\newblock In {\em ICML},  1139--1147.

\bibitem[\protect\citeauthoryear{Szegedy \bgroup et al\mbox.\egroup
  }{2015}]{szegedy2015going}
Szegedy, C.; Liu, W.; Jia, Y.; Sermanet, P.; Reed, S.; Anguelov, D.; Erhan, D.;
  Vanhoucke, V.; and Rabinovich, A.
\newblock 2015.
\newblock Going deeper with convolutions.
\newblock In {\em CVPR},  1--9.

\bibitem[\protect\citeauthoryear{Yim \bgroup et al\mbox.\egroup
  }{2017}]{yim2017gift}
Yim, J.; Joo, D.; Bae, J.; and Kim, J.
\newblock 2017.
\newblock A gift from knowledge distillation: Fast optimization, network
  minimization and transfer learning.
\newblock In {\em CVPR},  4133--4141.

\bibitem[\protect\citeauthoryear{Zagoruyko and
  Komodakis}{2016}]{zagoruyko2016wide}
Zagoruyko, S., and Komodakis, N.
\newblock 2016.
\newblock Wide residual networks.
\newblock In {\em BMVC},  87.1--87.12.

\bibitem[\protect\citeauthoryear{Zagoruyko and
  Komodakis}{2017}]{zagoruyko2016paying}
Zagoruyko, S., and Komodakis, N.
\newblock 2017.
\newblock Paying more attention to attention: Improving the performance of
  convolutional neural networks via attention transfer.
\newblock In {\em ICLR}.

\end{thebibliography}

\newpage
\input{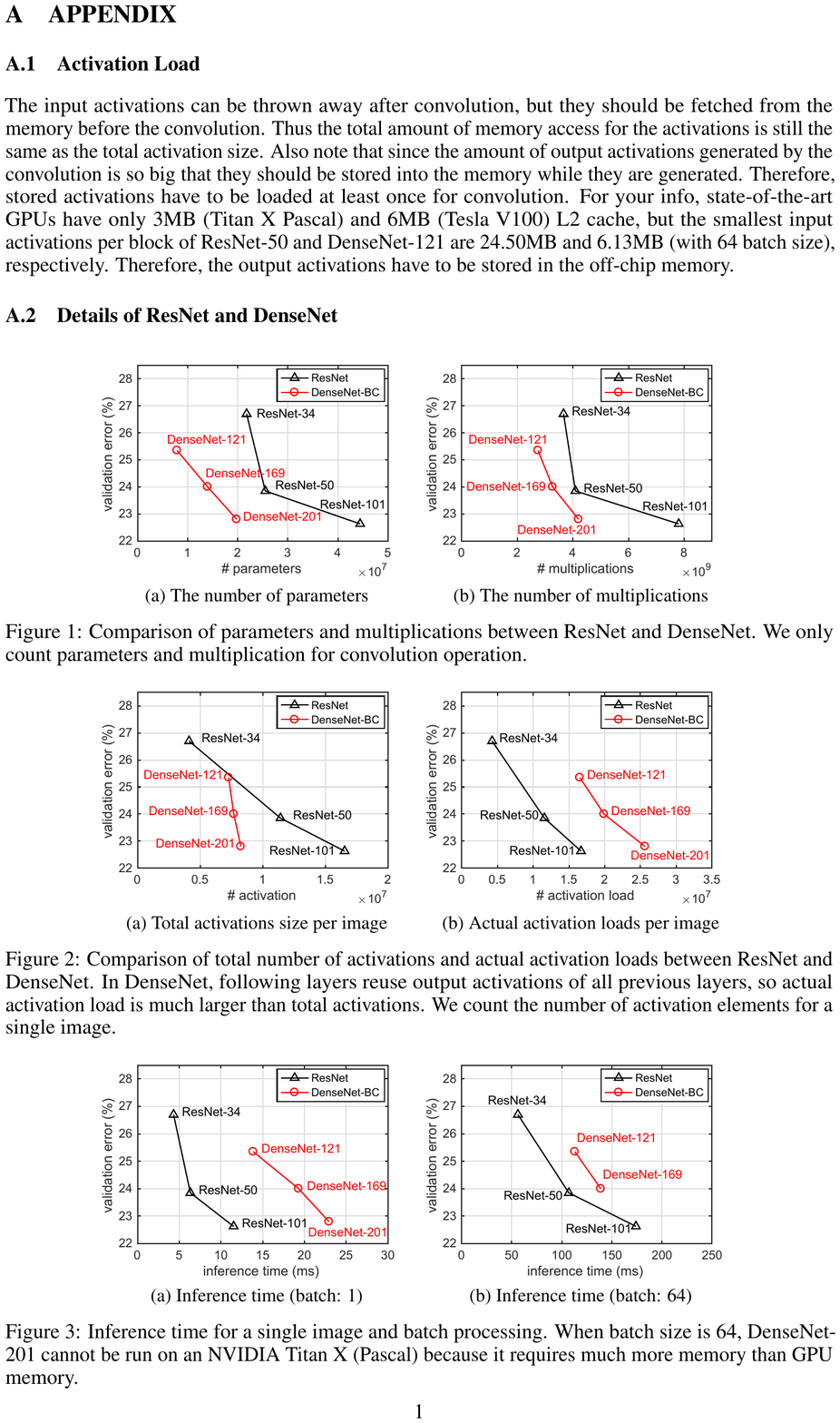}

\end{document}